\documentclass{fmcad}

\ifCLASSOPTIONcompsoc
  \usepackage[nocompress]{cite}
\else
  \usepackage{cite}
\fi
\usepackage{amsmath,amssymb,commath,amsthm}
\usepackage{algpseudocode}

\usepackage{subcaption} 
\usepackage{breakurl}

\usepackage{xspace}  
\newcommand{\name}{\mbox{\sc Art}\xspace}

\newcommand{\lone}{L_1}
\newcommand{\ltwo}{Euclid}

\newcommand{\absloss}{L}
\newcommand{\absdom}{\mathcal{D}}

\newcommand{\init}{\mathtt{in}}
\newcommand{\safe}{\mathtt{out}}

\newcommand{\para}[1]{\noindent{\bf\em #1}}

\newtheorem{theorem}{Theorem}[section]
\newtheorem{lemma}[theorem]{Lemma}

\theoremstyle{definition}
\newtheorem{definition}{Definition}[section]

\usepackage{booktabs}   
\usepackage{tikz}
\usepackage{pgfplots}
\usetikzlibrary{plotmarks}

\usepackage{bbold}  
\usepackage{bm}  

\usepackage[nounderscore]{syntax}  

\usepackage{soul}
\usepackage{makecell}
\usepackage{thmtools,thm-restate}

\begin{document}

%
\title{\name: Abstraction Refinement-Guided Training for\\Provably
  Correct Neural Networks}



\author{\IEEEauthorblockN{Xuankang Lin\IEEEauthorrefmark{1},
He Zhu\IEEEauthorrefmark{2},
Roopsha Samanta\IEEEauthorrefmark{1} and
Suresh Jagannathan\IEEEauthorrefmark{1}}
\IEEEauthorblockA{\IEEEauthorrefmark{1}Purdue University, West Lafayette, IN 47907}
\IEEEauthorblockA{\IEEEauthorrefmark{2}Rutgers University, Piscataway, NJ 08854}}


\maketitle

\begin{abstract}
  Artificial Neural Networks (ANNs) have demonstrated remarkable
  utility in various challenging machine learning applications. While
  formally verified properties of their behaviors are highly desired,
  they have proven notoriously difficult to derive and
  enforce. Existing approaches typically formulate this problem as a
  \emph{post facto} analysis process. In this paper, we present a
  novel learning framework that ensures such formal guarantees are
  \emph{enforced by construction}. Our technique enables training
  provably correct networks with respect to a broad class of safety
  properties, a capability that goes well-beyond existing approaches,
  \emph{without} compromising much accuracy.  Our key insight is that
  we can integrate an optimization-based abstraction refinement loop
  into the learning process and operate over dynamically constructed
  partitions of the input space that considers accuracy and safety
  objectives synergistically. The refinement procedure iteratively
  splits the input space from which training data is drawn, guided by
  the efficacy with which such partitions enable safety verification.
  We have implemented our approach in a tool (\name) and applied it to
  enforce general safety properties on unmanned aviator collision
  avoidance system ACAS Xu dataset and the Collision Detection
  dataset. Importantly, we empirically demonstrate that realizing
  safety does not come at the price of much accuracy. Our methodology
  demonstrates that an abstraction refinement methodology provides a
  meaningful pathway for building both accurate and correct machine
  learning networks.

\end{abstract}

%
\IEEEpeerreviewmaketitle

\section{Introduction}\label{sec:intro}


Artificial neural networks (ANNs) have emerged in recent years as the
primary computational structure for implementing many challenging
machine learning applications.  Their success has been due in large
measure to their sophisticated architecture, typically comprised of
multiple layers of connected neurons (or \emph{activation functions}),
in which each neuron represents a possibly non-linear function over
the inputs generated in a previous layer.  In a supervised setting,
the goal of learning is to identify the proper coefficients (i.e.,
\emph{weights}) of these functions that minimize differences between
the outputs generated by the network and ground truth, established via
training samples.  The ability of ANNs to identify fine-grained
distinctions among their inputs through the execution of this process
makes them particularly useful in a variety of diverse domains such as
classification, image recognition, natural language translation, or
autonomous driving.


However, the most {\em accurate} ANNs may still be {\em incorrect}.
Consider, for instance, the ACAS Xu (Airborne Collision Avoidance
System) application that targets avoidance of midair collisions
between commercial aircraft~\cite{Acas}, whose system is controlled by
a series of ANNs to produce horizontal maneuver advisories. One
example {\em safety property} states that if a potential intruder is
far away and is significantly slower than one's own vehicle, then
regardless of the intruder's and subject's direction, the ANN
controller should output a \textsf{Clear-of-Conflict} advisory (as it
is unlikely that the intruder can collide with the subject).
Unfortunately, even a sophisticated ANN handler used in the ACAS Xu
system, although well-trained, has been shown to violate this
property~\cite{Reluplex}.  Thus, ensuring the reliability of ANNs,
especially those adopted in safety-critical applications, is
increasingly viewed as a necessity.

The programming languages and formal methods community has responded
to this familiar, albeit challenging, problem with increasingly
sophisticated and scalable {\em verification}
approaches~\cite{Reluplex,ReluVal,AI2,AP+19} --- given a trained ANN
and a property, these approaches either certify that the ANN satisfies
the property or identify a potential violation of the property.
Unfortunately, when verification fails, these approaches provide no
insight on how to effectively leverage verification counterexamples to
{\em repair} complex, uninterpretable networks and ensure safety.
Further, many verification approaches focus on a popular, but
ultimately, narrow class of properties --- {\em local robustness} ---
expressed over {\em some, but not all} of a network's input space.

In this paper, we address the limitations of existing verification
approaches by proposing a novel {\em training} approach for
\emph{generation of ANNs that are correct-by-construction with respect
  to a broad class of correctness properties expressed over the
  network's inputs.} Our training approach integrates correctness
properties into the training objective through a {\em correctness loss
  function} that quantifies the violation of the correctness
properties. Further, to enable certification of correctness of a
possibly infinite set of network behaviors, our training approach
employs abstract interpretation methods~\cite{AI2,DiffAI} to generate
sound abstractions of both the \emph{input space} and the
\emph{network itself}.  Finally, to ensure the trained network is both
correct {\em and} accurate with respect to training data, our approach
iteratively refines the precision of the input abstraction, guided by
the value of the correctness loss function.  Our approach is sound ---
if the correctness loss reduces to $0$, the generated ANN is
guaranteed to satisfy the associated correctness properties.

\begin{figure}[t]
  \centering
  \includegraphics[width=0.9\columnwidth]{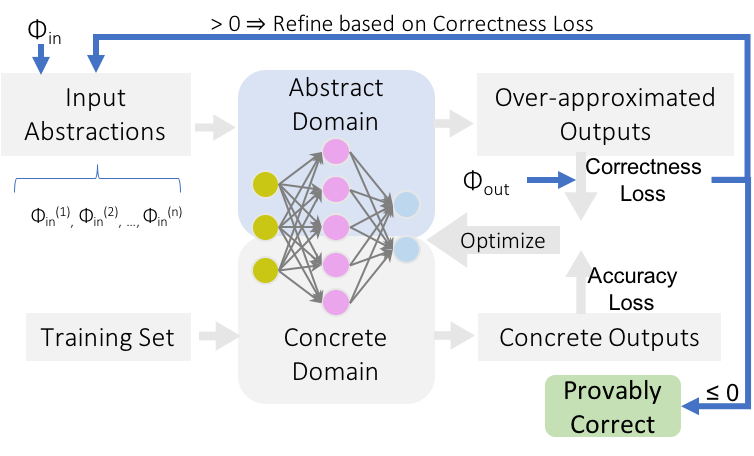}
\caption{The \name\ framework.}
\label{fig:vegalframework}
\end{figure}

The workflow of this overall approach --- Abstraction
Refinement-guided Training (\name) --- is shown in
Fig.~\ref{fig:vegalframework}.  \name takes as input a correctness
property $\left( \Phi_{\init}, \Phi_{\safe} \right)$ that prescribes
desired network output behavior using logic constraints $\Phi_{\safe}$
when the inputs to the network are within a domain described by
$\Phi_{\init}$.  \name is parameterized by an abstract domain
$\mathcal{D}$ that yields an abstraction over inputs in
$\Phi_{\init}$. Additionally, \name takes a set of labeled training
data. The correctness loss function quantifies the {\em distance} of
the abstract network output from the correctness constraint
$\Phi_{\safe}$.  In each training iteration, \name both updates the
network weights and refines the input abstraction.  The network
weights are updated using classical gradient descent optimization to
mitigate the correctness loss (upper loop of
Fig.~\ref{fig:vegalframework}) and the standard accuracy loss (lower
loop of Fig.~\ref{fig:vegalframework}).  The abstraction refinement
utilizes information provided by the correctness loss to improve
the precision of the abstract network output (the top arrow of
Fig.~\ref{fig:vegalframework}).
%
%
As we show in Section~\ref{sec:evaluation}, the key novelty of our
approach - exploiting the synergy between refinement and approximation
- (a) often leads to, at worst, \emph{mild} impact on accuracy
compared to a safe oracle baseline; and (b) provides significantly
higher assurance on network correctness than existing verification or
training~\cite{Gowal18} methods which do not exploit abstraction
refinement.

This paper makes the following contributions. (1) We present an
abstract interpretation-guided training strategy for building
correct-by-construction neural networks, defined with respect to a
rich class of safety properties, including functional correctness
properties that relate input and output structure. (2) We define an
input space abstraction refinement loop that reduces training on input
data to training on input space partitions, where the precision of the
abstraction is, in turn, guided by a notion of correctness loss as
determined by the correctness property. (3) We formalize soundness
claims that capture correctness guarantees provided by our
methodology; these results characterize the ability of our approach to
ensure correctness with respect to domain-specific correctness
properties. (4) We have implemented our ideas in a tool (\name) and
applied it to challenging benchmarks including the ACAS Xu collision
avoidance dataset~\cite{Acas,Reluplex} and the Collision Detection
dataset~\cite{Ehlers17}. We provide a detailed evaluation study
quantifying the effectiveness of our approach and assess its utility
to ensure correctness guarantees without compromising accuracy. We
additionally provide a comparison of our approach with \emph{post
  facto} counterexample-guided verification strategies to demonstrate
the benefits of \name's methodology compared to such techniques.

The remainder of the paper is organized as follows.  In the next
section, we provide a detailed motivating example that illustrates our
approach.  Section~\ref{sec:background} provides background and 
Section~\ref{sec:approach} formalizes our approach.
Details about \name's implementation and evaluation are provided in
Section~\ref{sec:evaluation}.  Related work and conclusions are presented
in Section~\ref{sec:related} and~\ref{sec:conc}, resp.


\section{Illustrative Example}
\label{sec:motivation}

\def\layersep{3cm}  

We illustrate and motivate the key components of our approach by
starting with a realistic, albeit simple, end-to-end example. We
consider the construction of a learning-enabled system for autonomous
driving.  The learning objective is to identify potentially dangerous
objects within a prescribed range of the vehicle's current position.\\


\para{Problem Setup.}  For the purpose of this example, we simplify
our scenario
by assuming that we track only a single object and that the
information given by the vehicle's radar is a feature vector of size
two, containing (a) the object's normalized relative speed
$v \in [-5, 5]$ where the positive values mean that the objects are
getting closer; and (b) the object's relative angular position
$\theta \in [-\pi, \pi]$ in a polar coordinate system with our vehicle
located in the center. Either action \textsf{Report} or action
\textsf{Ignore} is advised by the system for this object given the
information. 

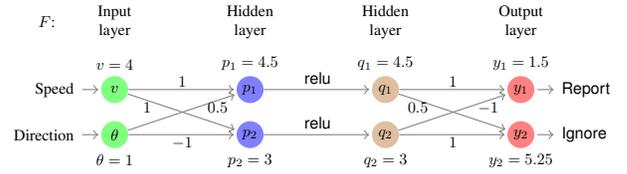
\begin{figure}[!t]
  \begin{tikzpicture}[shorten >=1pt,->,draw=black!50, node distance=\layersep,
    scale=0.6, every node/.style={transform shape}]
    \tikzstyle{every pin edge}=[<-,shorten <=1pt]
    \tikzstyle{neuron}=[circle,fill=black!25,minimum size=17pt,inner sep=0pt]
    \tikzstyle{input neuron}=[neuron, fill=green!50];
    \tikzstyle{output neuron}=[neuron, fill=red!50];
    \tikzstyle{hidden neuron}=[neuron, fill=blue!50];
    \tikzstyle{relu neuron}=[neuron, fill=brown!50];
    \tikzstyle{annot} = [text width=4em, text centered]

    \node (I-1) at (0, -1) [input neuron, pin=left:Speed, label={above:$v = 4$}] {$v$};
    \node (I-2) at (0, -2) [input neuron, pin=left:Direction, label={below:$\theta = 1$}] {$\theta$};

    \node (H-1) at (\layersep,-1) [hidden neuron, label={above:$p_1 = 4.5$}] {$p_1$};
    \node (H-2) at (\layersep,-2) [hidden neuron, label={below:$p_2 = 3$}] {$p_2$};

    \node (R-1) at (\layersep * 2, -1) [relu neuron, label={above:$q_1 = 4.5$}] {$q_1$};
    \node (R-2) at (\layersep * 2, -2) [relu neuron, label={below:$q_2 = 3$}] {$q_2$};

    \node (O-1) at (\layersep * 3, -1) [output neuron,pin={[pin
      edge={->}]right:\textsf{Report}}, label={above:$y_1 = 1.5$}] {$y_1$};
    \node (O-2) at (\layersep * 3, -2) [output neuron,pin={[pin
      edge={->}]right:\textsf{Ignore}}, label={below:$y_2 = 5.25$}] {$y_2$};

    \path[->]
    (I-1) edge node[sloped, yshift=5pt] {$1$} (H-1)
    (I-1) edge node[near start, xshift=-5pt, yshift=-3pt] {$1$} (H-2)
    (I-2) edge node[near end, xshift=5pt, yshift=-3pt] {$0.5$} (H-1)
    (I-2) edge node[sloped, yshift=-5pt] {$-1$} (H-2);

    \foreach \idx in {1,...,2}
    \path (H-\idx) edge node[above] {\textsf{relu}} (R-\idx);

    \path[->]
    (R-1) edge node[sloped, yshift=5pt] {$1$} (O-1)
    (R-1) edge node[near start, xshift=-5pt, yshift=-3pt] {$0.5$} (O-2)
    (R-2) edge node[near end, xshift=5pt, yshift=-3pt] {$-1$} (O-1)
    (R-2) edge node[sloped, yshift=-5pt] {$1$} (O-2);

    \node[annot,above of=I-1, node distance=1.5cm] (il) {Input\\layer};
    \node[annot,right of=il] (hl) {Hidden\\layer};
    \node[annot,right of=hl] (rl) {Hidden\\layer};
    \node[annot,right of=rl] {Output\\layer};
    \node[annot,left of=il, node distance=1.5cm] (func) {$F$:};
  \end{tikzpicture}
  \centering
  \caption{A monitoring system using 2-layer ReLU network.}
  \label{fig:demo-nn}
\end{figure}

Consider an implementation of an ANN for this problem that uses a
2-layer ReLU neural network $F$ with initialized weights as depicted
in Fig.~\ref{fig:demo-nn}. The network takes an input vector
$x = \left( v, \theta \right)$ and outputs a prediction score vector
$y = \left( y_1, y_2 \right)$ for actions \textsf{Report} and
\textsf{Ignore}, respectively. The action with higher prediction score
is picked by the advisory system. For simplicity, both layers in $F$
are linear layers with 2 neurons and without bias terms. An
element-wise ReLU activation function $relu(x) = \max(x, 0)$ is
applied after the first layer.\\

\para{Correctness Property.}  To serve as a useful advisory system, we
can ascribe some correctness properties that we would like the network
to always satisfy. While our approach generalizes to an arbitrary
number of the correctness properties that one may wish to enforce, we
focus on one such correctness property $\Phi$ in this example:
\emph{Objects in front of the vehicle that are stationary or moving
  closer should not be ignored}. The meaning of ``\emph{stationary or
  moving closer}'' and ``\emph{in front of}'' can be interpreted in
terms of predicates $\Phi_{\init}$ and $\Phi_{\safe}$ over feature
vector components such as $v \ge 0$ and
$\theta \in [0.5, 2.5]$\footnote{We pick $[0.5, 2.5]$ because it is
  slightly wider than the front view angle of
  $[\frac{\pi}{4}, \frac{3\pi}{4}]$.}, respectively.  Using such
representations and recalling that $v \in [-5,5]$,
$\Phi = \left(\Phi_{\init}, \Phi_{\safe}\right)$ can be precisely
formulated as:
\[
  \forall v, \theta. \; \underbrace{v \in [0, 5] \ \wedge \ \theta \in
  [0.5, 2.5]}_{\Phi_{\init}} \ \wedge \ y = F(v, \theta) \Rightarrow \underbrace{y_1 > y_2}_{\Phi_{\safe}}.
\]
Observe that this property is violated with the network and the
example input shown in Fig.~\ref{fig:demo-nn}.\\

\para{Concrete Correctness Loss Function.} To quantify how {\em
  correct} $F$ is on inputs satisfying predicate $\Phi_{\init}$, we
define a {\em correctness loss function}, denoted $\mathit{dist}_g$,
over the output $y$ of the neural network and the output predicate
$\Phi_{\safe}$:
\[\mathit{dist}_g(y,\Phi_{\safe}) = \min_{q \models \Phi_{\safe}} g(y,q),\]
parameterized on a distance function $g$ over the input space such as
the Manhattan distance ($\lone$-norm), Euclidean distance
($\ltwo$-norm), etc.  The correctness distance function is
intentionally defined to be semantically meaningful---when
$dist_g(y, \Phi_{\safe}) = 0$, it follows that $y$ satisfies the
output predicate $\Phi_{\safe}$. This function can then be used as a
loss function, among other training objectives to train the neural
network towards satisfying
$\left( \Phi_{\init}, \Phi_{\safe} \right)$.  For this example, we can
compute the correctness distance of the network output
$y = \left( y_1, y_2 \right)$ from $\Phi_{\safe} = y_1 > y_2$ to be
$dist_{\ltwo}(y, \Phi_{\safe}) = \max\left((y_2 - y_1) / \sqrt 2,
  0\right)$
which is calculated based on the Euclidean distance between point
$\left(y_1, y_2\right)$
and line $y_2 - y_1 = 0$.\\

\para{Abstract Domain.}  A general correctness property like $\Phi$ is
often defined over an infinite set of data points; however, since
training necessarily is performed using only a finite set of samples,
we cannot generalize observations made on just these samples to assert
the validity of $\Phi$ on the trained network. Our approach,
therefore, leverages abstract interpretation techniques to generate
sound abstractions of both the network input space and the network
itself. By training on these abstractions, our method obtains a finite
approximation of the infinite set of possible network behaviors,
enabling correct-by-construction training.

We parameterize our approach on any abstract domain that serves as a
sound over-approximation of a neural network's behavior, i.e.,
abstractions in which an abstract output is guaranteed to subsume all
possible outputs for the set of abstract inputs.  In the example, we
consider the \emph{interval} abstract domain $\mathcal I$
that
is simple enough to motivate the core ideas of our approach. We note
that \name is not bound to specific abstract domains, the interval
domain is used only for illustrative purposes here, our experiments in
Section~\ref{sec:evaluation} are conducted using more precise
abstractions.


\begin{figure}[!t]
  \begin{tikzpicture}[shorten >=1pt,->,draw=black!50, node distance=\layersep,
    scale=0.6, every node/.style={transform shape}]
    \tikzstyle{every pin edge}=[<-,shorten <=1pt]
    \tikzstyle{neuron}=[circle,fill=black!25,minimum size=17pt,inner sep=0pt]
    \tikzstyle{input neuron}=[neuron, fill=green!50];
    \tikzstyle{output neuron}=[neuron, fill=red!50];
    \tikzstyle{hidden neuron}=[neuron, fill=blue!50];
    \tikzstyle{relu neuron}=[neuron, fill=brown!50];
    \tikzstyle{annot} = [text width=4em, text centered]

    \node (I-1) at (0, -1) [input neuron, pin=left:Speed, label={above:$v \in [0, 5]$}] {$v$};
    \node (I-2) at (0, -2) [input neuron, pin=left:Direction, label={below:$\theta \in [0.5, 2.5]$}] {$\theta$};

    \node (H-1) at (\layersep,-1) [hidden neuron, label={above:$p_1 \in [0.25, 6.25]$}] {$p_1$};
    \node (H-2) at (\layersep,-2) [hidden neuron, label={below:$p_2 \in [-2.5, 4.5]$}] {$p_2$};

    \node (R-1) at (\layersep * 2, -1) [relu neuron, label={above:$q_1 \in [0.25, 6.25]$}] {$q_1$};
    \node (R-2) at (\layersep * 2, -2) [relu neuron, label={below:$q_2 \in [0, 4.5]$}] {$q_2$};

    \node (O-1) at (\layersep * 3, -1) [output neuron, pin={[pin
      edge={->}] right:\textsf{Report}}, label={above:$y_1 \in [-4.25, 6.25]$}] {$y_1$};
    \node (O-2) at (\layersep * 3, -2) [output neuron, pin={[pin
      edge={->}] right:\textsf{Ignore}}, label={below:$y_2 \in [0.125, 7.625]$}] {$y_2$};

    \path[->]
    (I-1) edge node[sloped, yshift=5pt] {$1$} (H-1)
    (I-1) edge node[near start, xshift=-5pt, yshift=-3pt] {$1$} (H-2)
    (I-2) edge node[near end, xshift=5pt, yshift=-3pt] {$0.5$} (H-1)
    (I-2) edge node[sloped, yshift=-5pt] {$-1$} (H-2);

    \foreach \idx in {1,...,2}
    \path (H-\idx) edge node[above] {\textsf{relu}} (R-\idx);

    \path[->]
    (R-1) edge node[sloped, yshift=5pt] {$1$} (O-1)
    (R-1) edge node[near start, xshift=-5pt, yshift=-3pt] {$0.5$} (O-2)
    (R-2) edge node[near end, xshift=5pt, yshift=-3pt] {$-1$} (O-1)
    (R-2) edge node[sloped, yshift=-5pt] {$1$} (O-2);

    \node[annot,above of=I-1, node distance=1.5cm] (il) {Input\\layer};
    \node[annot,right of=il] (hl) {Hidden\\layer};
    \node[annot,right of=hl] (rl) {Hidden\\layer};
    \node[annot,right of=rl] {Output\\layer};
    \node[annot,left of=il, node distance=1.5cm] (func) {$F_{\mathcal I}$:};
  \end{tikzpicture}
  \centering
  \caption{The 2-layer ReLU network over interval domain.}
  \label{fig:demo-nn-intvl}
\end{figure}
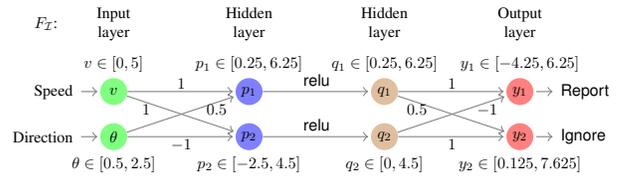

An interval abstraction of our 2-layer ReLU network, denoted
$F_{\mathcal I}$, is shown in Fig.~\ref{fig:demo-nn-intvl}. The
concrete neural network computation $F$ is abstracted by maintaining
the lower and upper bounds $[\underline u, \overline u]$ of each
neuron $u$. For neuron $p_2$ in this example, following interval
arithmetic \cite{IntvlMath}, the lower bound of neuron is computed by
$\underline{p_2} = 1 \cdot \underline v + (-1) \cdot \overline \theta
= -2.5$ and the upper bound
$\overline{p_2} = 1 \cdot \overline v + (-1) \cdot \underline \theta =
4.5$. For ReLU activation function, $F_{\mathcal I}$ resets negative
lower bounds to $0$ and preserves everything else. Consider neurons
$p_2 \rightarrow q_2$, lower bound $\underline{q_2}$ is reset to $0$
while its upper bound $\overline{q_2}$ remains unchanged. In this way,
$F_{\mathcal I}$ soundly over-approximates all possible outputs
generated by the network given any inputs satisfying
$\Phi_{\init}$. Applying $F_{\mathcal I}$, the neural network's
\emph{abstract output} is $y_1 \in [-4.25, 6.25]$ and
$y_2 \in [0.125, 7.625]$, which fails to show that $y_1 > y_2$ always
holds. As a counterexample depicted in Fig.~\ref{fig:demo-nn}, the
input $v=4 \land \theta =1$ leads to violation.\\

\para{Abstract Correctness Loss Function.}
Given $\Phi_{\init}$, to quantify how correct $F$ is based on the
abstract output $y^\#$, we can also define an abstract correctness
loss function, denoted $\absloss_g$, over $y^\#$ and the output
predicate $\Phi_{\safe}$:
\[
  \absloss_g(y^\#, \Phi_{\safe}) = \max_{y \in \gamma(y^\#)} dist_g(y, \Phi_{\safe}),
\]
where $\gamma(y^\#)$ maps $y^\#$ to the set of values it represents in
the concrete domain and $g$ is a distance function over the input
space as before. In our example,
$\absloss_{\ltwo}(y^\#, \Phi_{\safe}) = \max\left((\overline{y_2} -
  \underline{y_1}) / \sqrt 2, 0\right) = 11.875 / \sqrt 2$.

Measuring the worst-case distance of possible outputs to
$\Phi_{\safe}$, $\absloss_g$ is also semantically meaningful --- when
$\absloss_g(y^\#, \Phi_{\safe}) = 0$, it follows that all possible
values represented by $y^\#$ satisfy the output predicate
$\Phi_{\safe}$. In other words, the trained neural network $F$ is
certified safe w.r.t. the correctness property $\Phi$.

$\absloss_g$ can be leveraged as the objective function during
optimization. The $\min$ and $\max$ units in $\absloss_g$ can be
implemented using MaxPooling and MinPooling units, and hence is
differentiable. Then we can use off-the-shelf automatic
differentiation libraries~\cite{PyTorch} in the usual fashion to
derive and backpropagate the gradients and readjust $F$'s weights
towards minimizing $\absloss_g$.\\

\para{Input Space Abstraction Refinement.} The abstract correctness
loss function $\absloss_g$ provides a direction for neural network
weight optimization. However, $\absloss_g$ could be overly imprecise
since the amount of spurious cases introduced by the neural network
abstraction is correlated with the size of the abstract input {\em
  region}.
This kind of imprecision leads to sub-optimal optimization, ultimately
hurting the feasibility of correct-by-construction as well as the
model accuracy.

Such imprecision arises easily when using less precise abstract
domains like the interval domain. For our running example, by
bisecting the input space along each dimension, the resulting abstract
correctness loss values of each region range from $3.125 / \sqrt 2$ to
$9.125 / \sqrt 2$.
If the original abstract correctness loss $11.875 / \sqrt 2$ pertains
to a real input, it should be reflected in some sub-region as
well. Now that $9.125 / \sqrt 2 < 11.875 / \sqrt 2$, the original
abstract correctness loss must be spurious and thus suboptimal for
optimization.

To use more accurate gradients for network weight optimization, our
approach leverages the above observation to also iteratively partition
the input region $\Phi_{\init}$ during
training. 
In other words, we seek for an input space abstraction refinement
mechanism that reduces imprecise abstract correctness loss introduced
by abstract interpretation. Notably, incorporating input space
abstraction refinement with the gradient descent optimizer does not
compromise the soundness of our approach. As long as all sub-regions
of $\Phi_{\init}$ are provably correct, the network's
correctness with respect to $\Phi_{\init}$ trivially holds.\\


\para{Iterative Training.}
Our training algorithm interweaves input
space abstraction refinement and gradient descent training on a
network abstraction in each training iteration by leveraging the
correctness loss function produced by the network abstract interpreter
(as depicted in Fig.~\ref{fig:vegalframework}), until a provably
correct ANN is trained. The refined input abstractions computed in an iteration are used
for training over the abstract domain in the next iteration.

For our illustrative example, we set the learning rate of the
optimizer to be $0.01$.
In our experiment, the maximum correctness loss among all refined
input space abstractions drops to $0$ after 11 iterations.
Convergence was achieved by heuristically partitioning the input space
$\Phi_{\init}$ into 76 regions. The trained ANN is guaranteed to
satisfy the correctness property $(\Phi_{\init}, \Phi_{\safe})$.

\section{Background}
\label{sec:background}


\begin{definition}[Neural network]
  Neural networks are functions $F: \mathbb R^d \rightarrow \mathbb R^e$
  composed of $Q$ layers and $Q-1$ activation functions. Each layer is a
  function $f_k(\cdot) \in \mathbb R^{m_{k-1}} \rightarrow \mathbb R^{m_k}$ for
  $k = 1, \ldots, Q$ where $m_0 = d$ and $m_Q = e$. Each activation function is
  of the form $\sigma_k(\cdot) \in \mathbb R^{m_k} \rightarrow \mathbb R^{m_k}$ for
  $k = 1, \ldots, Q-1$. Then,
  $F = f_Q \circ \sigma_{Q-1} \circ f_{Q-1} \circ \ldots \circ \sigma_1 \circ
  f_1$.
\end{definition}

\begin{definition}[Abstraction]
  \label{def:absdom}
  An abstraction $\absdom$ is defined as a tuple:
  $\left< \absdom_c, \absdom_a, \alpha, \gamma, T \right>$ where
  \begin{itemize}
  \item $\absdom_c: \{ x ~ | ~ x \in \mathbb R^d \}$ and where
    $d \in \mathbb Z^+$ is the concrete domain;
  \item $\absdom_a$ is the abstract domain of interest;
  \item $\alpha(\cdot)$ is an \emph{abstraction} function that maps a
    set of concrete elements to an abstract element;
  \item $\gamma(\cdot)$ is a \emph{concretization} function that maps 
    an abstract element to a set of concrete elements;
  \item
    $T = \left\{ \left( T_c, T_a \right) ~\vert~ T_c(\cdot): \absdom_c \rightarrow 
      \absdom_c, T_a(\cdot): \absdom_a
      \rightarrow \absdom_a \right\}$ is a set of transformer
    pairs over $\absdom_c$ and $\absdom_a$.
  \end{itemize}
  An abstraction is sound if for all $S \subseteq \absdom_c$,
  $S \subseteq \gamma(\alpha(S))$ holds and given
  $\left( T_c, T_a \right) \in T$,
  \[
    \forall c \in \absdom_c, a \in \absdom_a, \ c \in \gamma(a)
    \implies T_c(c) \in \gamma(T_a(a)).
  \]
\end{definition}

\begin{definition}[$\absdom$-compatible]
  \label{def:d-compatible}
  Given a sound abstraction
  $\absdom = \left< \absdom_c, \absdom_a, \alpha, \gamma, T
  \right>$, a neural network $F$ is $\absdom$-compatible iff for
  every layer or activation function $\iota(\cdot)$ in $F$, there exists
  an abstract transformer $T_a$ such that
  $\left(\iota(\cdot), T_a \right) \in T$, and $T_a$ is differentiable at
  least almost everywhere.

  For a $\absdom$-compatible neural network $F$, we denote by
  $F_{\absdom}: \absdom_a \rightarrow \absdom_a$ the
  over-approximation of $F$ where every layer $f_k(\cdot)$ and
  activation function $\sigma_k(\cdot)$ in $F$ are replaced in
  $F_{\absdom}$ by their corresponding abstract transformers in
  $\absdom$.
\end{definition}

Although our approach is parametric over abstract domains, we do
require every abstract transformer $T_a$ associated with these domains
to be differentiable, so as to enable training using the worst cases
over-approximated over $\absdom$ via gradient-descent style
optimization algorithms.

To reason about a neural network over an abstraction $\absdom$, we need to
first characterize what it means for an ANN to operate over $\absdom$.
\begin{definition}[Evaluation over Abstract Domain]
  \label{def:approx-eval}
  Given a $\absdom$-compatible neural network $F$,
  the evaluation of $F$ over $\absdom$ and a
  range of inputs $X \in \absdom_a$ is $F_{\absdom}(X)$ where
  $F_{\absdom}(X)$ over-approximates all possible outputs in
  the concrete domain corresponding to any input covered by $X$.  
\end{definition}

\begin{restatable}[Over-approximation Soundness]{theorem}{overapproxsoundness}
  \label{thm:approx-soundness}
  For sound abstraction $\absdom$, given a $\absdom$-compatible
  neural network $F$, a range of inputs $X \in \absdom_a$,
  \[
    \forall x. \; {x} \in \gamma(X) \, \Rightarrow \, F\left( x
    \right) \in \gamma\left( F_{\absdom}\left( X
      \right)\right).
  \]
\end{restatable}

Proofs of all theorems are provided in the Appendix.

\section{Correct-by-Construction Training}
\label{sec:approach}

Our approach aims to train an ANN $F$ with respect to a
\emph{correctness property} $\Phi$, which is formally defined in
Section~\ref{subsec:safety-prop}.
The abstraction of $F$ w.r.t. $\Phi$ based on abstract domain
$\absdom$ essentially can be seen as a function parameterized over the
weights of $F$, which can nonetheless be trained to fit $\Phi$ using
standard optimization algorithms.
Section~\ref{subsec:safety-loss} formally defines the
abstract \emph{correctness loss} function $\absloss_{\absdom}$ to
guide the optimization of $F$'s weights over $\absdom$.
Such an abstraction inevitably introduces spurious data samples into
training due to over-approximation.
%
%
Section~\ref{subsec:absref} introduces the idea of \emph{input
  space abstraction and refinement} as a mechanism that can reduce
such spuriousness during optimization over $\absdom$.
%
%
The detailed pseudocode of \name algorithm, including the refinement
procedure, is presented in Section~\ref{subsec:art-alg}.

\subsection{Correctness Property}
\label{subsec:safety-prop}

The correctness properties we consider are expressed as logical
propositions over the network's inputs and outputs.   
We assume that an ANN correctness property expresses constraints on the
outputs, given assumptions on the inputs.

\begin{definition}[Correctness Property]
  \label{def:safety-prop}
  Given a neural network $F: \mathbb R^d \rightarrow \mathbb R^e$, a
  correctness property
  $\Phi = \left( \Phi_{\init}, \Phi_{\safe} \right)$ is a tuple in
  which $\Phi_{\init}$ defines a bounded input domain over
  $\mathbb R^d$ in the form of an interval
  $[\underline{x}, \overline{x}]$ where
  $\underline{x}, \overline{x} \in \mathbb R^d$, are lower, upper
  bounds, resp., on the network input; and $\Phi_{\safe}$ is a
  quantifier-free Boolean combination of linear inequalities over the
  network output vector $y \in \mathbb R^e$: 
  \begin{itemize}
  \item[]
    \begin{grammar}
      <$\Phi_{\safe}$> ::= <P> | $\neg$<P> | <P> $\land$ <P> | <P> $\lor$ <P>;

      <P> ::= $A \cdot y \le b$ where $A \in \mathbb R^e, \, b \in \mathbb R$;
    \end{grammar}
  \end{itemize}
\end{definition}

An input vector $x \in \mathbb R^d$ is said to satisfy
$\Phi_{\init}=[\underline{x}, \overline{x}]$, denoted
$x \models \Phi_{\init}$, iff
$\underline{x} \leq x \leq \overline{x}$.  An output vector
$y \in \mathbb R^e$ satisfies $\Phi_{\safe}$, denoted
$y \models \Phi_{\safe}$, iff $\Phi_{\safe}(y)$ is true. A neural
network $F: \mathbb R^d \rightarrow \mathbb R^e$ satisfies $\Phi$,
denoted $F \models \Phi$, iff
$\forall x. \; x \models \Phi_{\init} \, \Rightarrow \, F(x) \models
\Phi_{\safe}$.


\begin{definition}[Concrete Correctness Loss Function]
  \label{def:conc-loss}
  For an atomic output predicate $P$, the concrete correctness loss
  function, $dist_g(y, P)$, quantifies the {\em distance} from an
  output vector $y \in \mathbb R^e$ to $P$:
  \[
    dist_g(y, P) = \min_{q \models P} g(y, q)
  \]
  where $g: \mathbb R^d \times \mathbb R^d \mapsto \mathbb Z^{\geq 0}$ 
  is a differentiable distance function over the inputs.
  Similarly, $dist_g(y, \Phi_{\safe})$, the ``\emph{distance}'' from
  an output vector $y \in \mathbb R^e$ to general output predicate
  $\Phi_{\safe}$, can be computed efficiently by induction as long as
  $g(\cdot,\cdot)$ can be computed efficiently:
  \begin{itemize}
  \item $dist_g(y, P)$ and $dist_g(y, \neg P)$ can be computed using basic
    arithmetic;
  \item $dist_g(y, P_1 \land P_2)$ = $\max(dist_g(y, P_1), dist_g(y, P_2))$;
  \item $dist_g(y, P_1 \lor P_2)$ = $\min(dist_g(y, P_1), dist_g(y, P_2))$.
  \end{itemize}
\end{definition}
Note that $dist_g(y, \Phi_{\safe})$ may not represent the minimum
distance for arbitrary $\Phi_{\safe}$, but it is efficient to compute
while still retaining the following soundness theorem.

\begin{restatable}[Zero Concrete Correctness Loss Soundness]{theorem}{concretelosssoundness}
  \label{thm:concrete-loss-soundness}
  Given output predicate $\Phi_{\safe}$ over $\mathbb R^e$ and output
  vector $y \in \mathbb R^e$,
  \[
    dist_g(y, \Phi_{\safe}) = 0 \implies y \models \Phi_{\safe}.
  \]
\end{restatable}


\subsection{Over-approximation}
\label{subsec:safety-loss}

To reason about correctness properties defined over an infinite set of
data points, our approach generates sound abstractions of both the
network input space and the network itself, obtaining a finite
approximation of the infinite set of possible network behaviors. We
start by quantifying the abstract correctness loss of
over-approximated outputs.

\begin{definition}[Abstract Correctness Loss Function]
  \label{def:safety-dist}
  Given a sound abstraction
  $\absdom = \left< \absdom_c, \absdom_a, \alpha, \gamma, T \right>$,
  a $\absdom$-compatible neural network $F$, and a correctness
  property $\Phi = \left( \Phi_{\init}, \Phi_{\safe} \right)$, the
  abstract correctness loss function is defined as:
  \[
    \begin{split}
      \absloss_{\absdom,g}(F, \Phi) &= \max_{p \in
        \gamma(Y_{\absdom})} dist_g(p, \Phi_{\safe})\\
      \text{where } Y_{\absdom} &= F_{\absdom}(\alpha(\Phi_{\init})).
    \end{split}
  \]
  Here $g: \mathbb R^d \times \mathbb R^d \mapsto \mathbb Z^{\geq 0}$
  is a differentiable distance function over concrete inputs as
  before. 
\end{definition}
The abstract correctness loss function measures the worst-case
distance to $\Phi_{\safe}$ of any neural network outputs subsumed by
the abstract network output.
%
%
It is designed to extend the notion of concrete correctness loss to
the abstract domain with a similar soundness guarantee, as formulated
in the following theorem.
\begin{restatable}[Zero Abstract Correctness Loss Soundness]{theorem}{zerodist}
  \label{thm:zero-dist}
  Given a sound abstraction $\absdom$, a $\absdom$-compatible
  neural network $F$, and a correctness property $\Phi$,
  \[
    \absloss_{\absdom,g}(F, \Phi) = 0 \implies F \models \Phi.
  \]
\end{restatable}

In what follows, we fix the distance function $g$ over concrete inputs 
and denote the abstract correctness loss function simply as $\absloss_\absdom$.

\subsection{Abstraction Refinement}
\label{subsec:absref}

Recall that in Section~\ref{sec:motivation} we illustrated how
imprecision in the correctness loss for a coarse abstraction
can be mitigated using an input space abstraction refinement
mechanism.  Our notion of refinement is formally defined below.

\begin{definition}[Input Space Abstraction]
  \label{def:splitting}
  An input space abstraction $S$ refines a correctness property
  $\Phi = \left(\Phi_{\init}, \Phi_{\safe} \right)$ into a set of
  correctness properties
  $S = \left\{\left(\Phi^i_{\init}, \Phi_{\safe} \right)\right\}$ such
  that $\Phi_{\init} = \bigcup_i \Phi^i_{\init}$. Given a neural
  network $F$ and an input space abstraction $S$,
  $F \models S \iff \bigwedge_{\Phi \in S} F \models \Phi$.
\end{definition}

\begin{definition}[Input Space Abstraction Refinement]
  \label{def:splitting-refinement}
  A well-founded abstraction refinement $\sqsubseteq$ is a binary
  relation over a set of input abstractions
  $\mathcal{S} = \{S_1, S_2, \ldots\}$ such that:
  \begin{itemize}
  \item (reflexivity): $\forall S_i \in \mathcal{S}$, $\, S_i \sqsubseteq S_i$;
  \item (refinement): $\forall S_i \in \mathcal{S}$ and correctness
    property
    $(\Psi_{\init}, \Psi_{\safe})$,
    \begin{equation*}
      \begin{split}
        &\left(\Psi_{\init} = \bigcup_{(\Phi^j_{\init}, \_) \in S_i} \Phi^j_{\init}\right)
        \land
        \left(\bigwedge_{(\_, \Phi^j_{\safe}) \in S_i} \Phi^j_{\safe} \Leftrightarrow \Psi_{\safe}\right)
        \\
        &\implies S_i \sqsubseteq \{(\Psi_{\init}, \Psi_{\safe})\};
      \end{split}
    \]

  \item (transitivity): $\forall S_1, S_2, S_3 \in \mathcal{S}$,
    $S_1 \sqsubseteq S_2 \land S_2 \sqsubseteq S_3 \implies S_1
    \sqsubseteq S_3$;

  \item (composition):
    $\forall S_1, S_2, S_3, S_4 \in \mathcal{S}, S_1 \sqsubseteq S_3 \land S_2
    \sqsubseteq S_4 \implies S_1 \cup S_2 \sqsubseteq S_3 \cup S_4$.
  \end{itemize}
\end{definition}

The \emph{reflexivity}, \emph{transitivity}, and \emph{compositional}
requirements for a well-founded refinement are natural. The
\emph{refinement} rule states that an input space abstraction $S$
refines some correctness property
$\left( \Psi_{\init}, \Psi_{\safe} \right)$ if the union of all input
domains in $S$ is equivalent to $\Psi_{\init}$ and all output
predicates in $S$ are logically equivalent to $\Psi_{\safe}$. This
rule enables $\Psi_{\init}$ to be safely decomposed into a set of
sub-domains. As a result, the problem of enforcing coarse-grained
correctness properties on neural networks can be converted into one
that enforces multiple fine-grained properties, an easier problem to
tackle because much of the imprecision introduced by the
coarse-grained abstraction can now be eliminated.
%
%
\begin{restatable}[Sufficient Condition via Refinement]{theorem}{sufficientviarefine}
  \label{thm:sufficient-via-refine}
  \[
    \forall F, S_1, S_2, ~ S_1 \sqsubseteq S_2 \land F \models S_1
    \implies F \models S_2.
  \]
\end{restatable}
\noindent To do this, we naturally extend the notion of abstract
correctness loss over one property to an input space abstraction.
\begin{definition}[Abstract Correctness Loss Function for Input Space
  Abstraction]
  \label{def:safety-loss}
  Given a sound abstraction $\absdom$, $\absdom$-compatible neural
  network $F$, and input space abstraction $S$, the \emph{abstract
    correctness loss} of $F$ with respect to $S$ is denoted
  by\footnote{We can refine the definition to have positive weighted
    importance of each correctness property in $S$; ascribing
    different weights to different correctness properties does not
    affect soundness.}
  \[
    \absloss_{\absdom}(F, S) = \sum_{\Phi \in S} \absloss_{\absdom}(F, \Phi).
  \]
\end{definition}

\begin{restatable}[Zero Abstract Correctness Loss for Input Space
  Abstraction]{theorem}{zerosafeloss}
  \label{thm:zero-safe-loss}
  Given a sound abstraction $\absdom$, a $\absdom$-compatible neural
  network $F$, and an input space abstraction $S$,
  \[
    \absloss_{\absdom}(F, S) = 0 \implies F \models S.
  \]
\end{restatable}

\subsection{The \name Algorithm}
\label{subsec:art-alg}

\begin{figure}[!t]
  \caption{\name correct-by-construction training algorithm.}
  \label{alg:vegal}
  \begin{algorithmic}[1]
    \Require Abstract domain $\absdom$, $\absdom$-compatible neural
    network $F$, input space abstraction $S$, learning rate
    $\eta \in \mathbb R^+$, training data set
    $\{(x_\text{train}, y_\text{label})\}$, accuracy loss function
    $L_{\mathcal A}$,
    accuracy loss bound $\epsilon_{\mathcal A} \in \mathbb R^+$,
    hyper-parameter $k$.
    \Ensure Return the optimized $F$ whose correctness properties are
    enforced and accuracy loss bounded by $\epsilon_{\mathcal A}$.
    \Statex

    \Procedure{\name}{}
    \State $\vec W \gets$ all weights in $F$ to optimize
    \While{{\tt true}}  \label{alg:vegal:startwhile}
    \State \label{alg:vegal:check-start} $\ell_{\absdom}, \ \ell_{\mathcal A} \gets \absloss_{\absdom}(F, S), \ L_{\mathcal A}(F, \{(x_\text{train}, y_\text{label})\})$
    \If{$\ell_{\absdom} = 0\ \wedge\ \ell_{\mathcal A}  \le \epsilon_{\mathcal A}$}
    \State \Return $F$\;
    \EndIf
    \label{alg:vegal:check-end}

    \Statex
    \Comment{optimization}
    \State \label{alg:vegal:optimize-start}
    $\nabla F \gets \frac{\partial (\ell_{\absdom} +
      \ell_{\mathcal A})}{\partial \vec W}$\;
    \State $\vec W \gets \vec W - \eta \cdot \nabla F$\;
    \label{alg:vegal:optimize-end}

    \Statex
    \Comment{refinement}
    \State $T \gets$ Subset of $S$ with $k$ largest $\ell_{\absdom}$ values\label{alg:vegal:refine-start} 
    \State $S' \gets S \setminus T$
      \ForAll{$(\Phi^i_{\init}, \Phi^i_{\safe}) \in T$} \label{alg:vegal:prop-occur-1}
        \ForAll{$\Psi^j_{\init} \in$ \Call{Refine}{$\Phi^i_{\init}$, $\ell_{\absdom}$}}
        \State $S' \gets S' \cup \{ (\Psi^j_{\init}, \Phi^i_{\safe}) \}$\;
        \EndFor \label{alg:vegal:prop-occur-2}
      \EndFor
      \State $S \gets S'$\;
      \label{alg:vegal:refine-end}
    \EndWhile
    \label{alg:vegal:endwhile}
    \EndProcedure

    \Statex
    \Procedure{Refine}{$\Psi_{\init}$, $\ell_{\absdom}$}
    \ForAll{dimension $i$ of $\Psi_{\init}$}
      \State score$_i = \frac{\partial \ell_{\absdom}}{\partial \{\Psi_{\init}\}_i} \times |\{\Psi_{\init}\}_i|$
    \EndFor
    \State $dim \gets \arg\max \text{score}_i$
    \Comment{pick dimension}
    \State $\Psi_{\init}^1, \Psi_{\init}^2 \gets \Psi_{\init}$ bisected along dimension $dim$
    \State \Return $\left\{ \Psi_{\init}^1, \Psi_{\init}^2 \right\}$
    \EndProcedure
  \end{algorithmic}
\end{figure}

The goal of our ANN training algorithm, given in Fig.~\ref{alg:vegal},
is to optimize the network to have $L_{\absdom}(F, S)$ reduce to $0$,
thereby ensuring a correct-by-construction network. The algorithm
takes as input both an initial input space abstraction $S$ and a set
of labeled training data
$\left\{ \left( x_\text{train}, y_\text{label} \right) \right\}$ in
order to achieve correctness while maintaining high accuracy on the
trained model. The abstract correctness loss, denoted
$\ell_{\absdom}$, is computed at Line~\ref{alg:vegal:check-start}
according to Def.~\ref{def:safety-dist} and 
checked correctness by comparing against $0$. If $\ell_{\absdom} = 0$,
as long as the accuracy loss, denoted $\ell_{\mathcal A}$, is also
satisfactory, \name returns a correct and accurate network following
Thm.~\ref{thm:zero-safe-loss}.

The joint loss of $\ell_{\absdom}$ and $\ell_{\mathcal A}$ is used to
guide the optimization of neural network parameters using standard
gradient-descent algorithms. The requirement of abstract transformers
being differentiable at least almost anywhere in
Def.~\ref{def:d-compatible} enables computation of gradients
$\ell_{\absdom}$ using off-the-shelf automatic differentiation
libraries~\cite{PyTorch}.

Starting from Line~\ref{alg:vegal:refine-start}, abstractions in $S$
that have the largest $\ell_{\absdom}$ values represent the
potentially most imprecise cases and thus are chosen for
refinement. During refinement, \name first picks a dimension to refine
using heuristic scores similar to \cite{ReluVal}. The heuristic
coarsely approximates the cumulative gradient over one dimension, with
a larger score suggesting greater potential of decreasing correctness
loss. The input abstraction is then bisected
along the picked dimension as refinement.

\begin{restatable}[\name Soundness]{corollary}{artsoundness}
  Given a sound abstraction $\absdom$, a $\absdom$-compatible neural
  network $F$, and an initial input space abstraction $S$ of
  correctness properties, if the \name algorithm in
  Fig.~\ref{alg:vegal} generates a neural network $F'$,
  $\absloss_{\absdom}(F',S)=0$ and $F' \models S$.
\end{restatable}


\section{Evaluation}
\label{sec:evaluation}

We have performed an evaluation of our approach to validate the
feasibility of building neural networks that are
correct-by-construction over a range of correctness
properties.\footnote{The code is available at
  \url{https://github.com/XuankangLin/ART}.} All experiments reported
in this section were performed on a Ubuntu 16.04 system with 3.2GHz
CPU and NVidia GTX 1080 Ti GPU with 11GB memory. All experiments uses
the DeepPoly abstract domain \cite{DeepPoly} implemented on Python 3.7
and PyTorch 1.4~\cite{PyTorch}.

\subsection{ACAS Xu Dataset}
Our first evaluation study centers around the network architecture and
correctness properties described in the Airborne Collision Avoidance
System for Unmanned Aircraft (ACAS Xu) dataset~\cite{Acas,Reluplex}. A
family of $45$ neural networks are used in the avoidance system; each
of these networks consists of $6$ hidden layers with $50$ neurons in
each hidden layer. ReLU activation functions are applied to all hidden
layer neurons.  All $45$ networks take a feature vector of size $5$ as
input that encodes various aspects of an airborne environment. The
outputs of the networks are prediction scores over $5$ advisory
actions to select the advisory action.

In the evaluation, we reason about sophisticated correctness
conditions of the ACAS Xu system in terms of its aggregated ability to
preserve up to $10$ correctness properties~\cite{Reluplex} among all
$45$ networks. Each network is supposed to satisfy some subset of
these $10$ properties. All correctness properties $\Phi$ can be
formulated in terms of input ($\Phi_{\init}$) and output
($\Phi_{\safe}$) predicates as in Section \ref{subsec:safety-prop}.\\

\para{Setup.} \label{subsec:acas-results}
Among the $45$ provided networks, $36$ are reported with safety
property violations and $9$ are reported safe~\cite{Reluplex}. We
evaluate \name on those $36$ unsafe networks to demonstrate the
effectiveness of generating correct-by-construction networks. The test
sets from unsafe networks may contain unsafe points and are thus
unauthentic, so we apply \name on those $9$ already safe networks to
demonstrate the accuracy overhead when enforcing the safety
properties. Unfortunately, the training and test sets to build these
ACAS Xu networks are not publicly available online. In spite of that,
the ACAS Xu dataset provides the state space of input states that is
used for training and over which the correctness properties are
defined. We, therefore, uniformly sample a total of $10$k training set
and $5$k test set data points from the state space. The labels are
collected by evaluating each of the provided $45$ networks on these
sampled inputs, with those ACAS Xu networks serving as oracles. Each
network is then trained by \name using its safety specification and
the prepared training set, starting with the provided weights when
available or otherwise randomly initialized weights. We record whether
the trained network is correct-by-construction, as well as their
accuracy evaluated on
the prepared test set and the overall training time.\\

\para{Applying \name.}
During each training epoch (i.e., each iteration of the outermost
while loop in Fig.~\ref{alg:vegal}), our implementation refines up to
$k=200$ abstractions at a time that expose the largest correctness
losses. Larger $k$ leads to finer-grained abstractions but incurs more
training cost. The Adam optimizer \cite{Adam} is used in both training
tasks and runs up to $100$ epochs with learning rate $0.001$ and a
learning rate decay policy if the loss has been stable for some
time. Cross entropy loss is used as the loss function for accuracy
. For all experiments with refinement enabled, refinement operations
are applied to derive up to $5$k refined input space abstractions
before weight update starts. The detailed results are shown in
Table~\ref{tab:acas-numbers}.

\begin{table}[!t]
  \caption{Applying \name\ to ACAS Xu Dataset.}
  \label{tab:acas-numbers}
  \centering
  \setlength{\tabcolsep}{3pt}
  \resizebox{\columnwidth}{!}{
  \begin{tabular}{l *6c}
    \toprule
    {} & Refinement & Safe\% & Min Accu. & Mean Accu. & Max Accu.\\
    \midrule
    $36$ unsafe nets & \makecell{Yes\\No} & \makecell{\textbf{100\%}\\$94.44\%$} & \makecell{$90.38\%$\\$87.88\%$} & \makecell{$96.10\%$\\$94.45\%$} & \makecell{$98.70\%$\\$98.22\%$}
    \\
    \addlinespace
    $9$ safe nets & \makecell{Yes\\No} & \makecell{\textbf{100\%}\\$88.89\%$} & \makecell{$93.82 \%$\\$86.32\%$} & \makecell{$96.25\%$\\$94.29\%$} & \makecell{$99.92\%$\\$99.92\%$}\\
    \bottomrule
  \end{tabular}
  }
\end{table}

To demonstrate the importance of abstraction refinement mechanism, we
also compare between the results with and without refinement (as done
in existing work \cite{DiffAI}).  For completeness, we record the
correct-by-construction enforced rate (\textsf{Safe\%}) and the
evaluated accuracy statistics for both tasks among multiple
runs. Observe that \name successfully generates
correct-by-construction networks for all scenarios with only minimal
loss in accuracy. On the other hand, if refinement is disabled, it
fails to generate correct-by-construction networks for all cases, and
displays lower accuracy than the refinement-enabled
instantiations. The average training time for each network is 69.39s
if with refinement and 57.85s if without.\\

\begin{figure}[!t]
  \centering
  \begin{minipage}[t]{\columnwidth}
    \begin{subfigure}[t]{0.48\columnwidth}
      \begin{tikzpicture}[scale=0.55]
        \begin{axis}[
          title=Correctness rate (\%),
          x tick label style={/pgf/number format/1000 sep=},
          xtick=data,
          symbolic x coords={$N_1$, $N_2$, $N_3$, $N_4$, $N_5$, $N_6$, $N_7$, $N_8$},
          legend style={at={(0.85, 0.95)}, anchor=north east},
          ybar,
          ymin=-1, ymax=110,
          bar width=7pt
          ]
          \addplot coordinates {
            ($N_1$, 20)
            ($N_2$, 20)
            ($N_3$, 100)
            ($N_4$, 0)
            ($N_5$, 0)
            ($N_6$, 0)
            ($N_7$, 0)
            ($N_8$, 100)
          };
          \addplot coordinates {
            ($N_1$, 80)
            ($N_2$, 60)
            ($N_3$, 100)
            ($N_4$, 0)
            ($N_5$, 40)
            ($N_6$, 40)
            ($N_7$, 0)
            ($N_8$, 100)
          };
          \legend{Sampled points, Counterexamples}  
        \end{axis}
      \end{tikzpicture}
      \centering
    \end{subfigure}
    ~
    \begin{subfigure}[t]{0.48\columnwidth}
      \begin{tikzpicture}[scale=0.55]
        \begin{axis}[
          title=Accuracy change (\%),
          x tick label style={/pgf/number format/1000 sep=},
          xtick=data,
          yticklabel pos=right,
          symbolic x coords={0, $N_1$, $N_2$, $N_3$, $N_4$, $N_5$, $N_6$, $N_7$, $N_8$},
          enlargelimits=0.15,
          legend style={at={(0.95, 0.95)}, anchor=north east},
          ybar,
          bar width=7pt,
          after end axis/.append code={
            \draw ({rel axis cs:0,0}|-{axis cs:0,0}) -- ({rel axis cs:1,0}|-{axis cs:0,0});
          }
          ]
          \addplot coordinates {
            ($N_1$, 13.40)
            ($N_2$, -1.43)
            ($N_3$, 4.61)
            ($N_4$, 3.20)
            ($N_5$, -2.45)
            ($N_6$, -0.74)
            ($N_7$, -2.55)
            ($N_8$, 0.)
          };
          \addplot coordinates {
            ($N_1$, 5.41)
            ($N_2$, -20.88)
            ($N_3$, -5.05)
            ($N_4$, -6.64)
            ($N_5$, -15.67)
            ($N_6$, -21.52)
            ($N_7$, -6.99)
            ($N_8$, 0.)
          };
          \legend{Sampled points, Counterexamples}
        \end{axis}
      \end{tikzpicture}
      \centering
    \end{subfigure}
  \end{minipage}
  \caption{Correctness rate and accuracy change of \emph{post facto}
    training using sampled points or counterexamples. Results are
    normalized based on the baseline networks.}
  \label{fig:eval-conc-pts}
\end{figure}
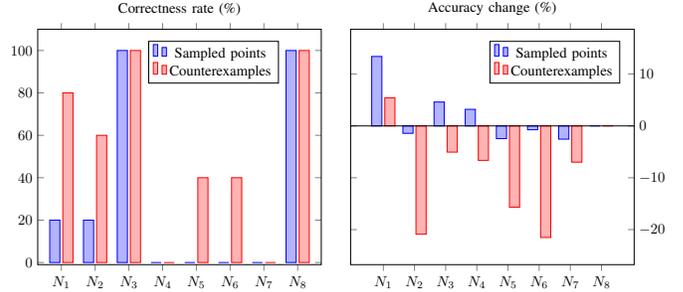

\para{Comparison with \emph{post facto} training loop.}
We also consider a comparison of our abstraction refinement-guided
training for correct-by-construction networks against a \emph{post
  facto} training loop that feed concrete correctness related data
points to training loops. Such concrete points may be sampled from the
provided specification or the collected counterexamples from an external
solver. We show the results on $8$ representative networks comparing
to the same baseline in Figure~\ref{fig:eval-conc-pts}. These $8$
networks belong to a representative set of networks that cover all
$10$ provided safety properties.

For the experiment using sampled data points, $5$k points sampled from
correctness properties are used during training. For the experiment
using counterexamples, all counterexamples from correctness queries to
external verifier \textsf{ReluVal}~\cite{ReluVal} are collected and
used during training. In both experiments, the points from original
training set are used for jointly training to preserve accuracy and
the correctness distance functions following that in
Section~\ref{subsec:safety-loss} are used as loss functions. We
concluded the experiments using counterexamples after $20$ epochs
since no improvement was seen after this point. Both experiments fail
to enforce correctness properties in most cases and they may impose
great impact to model accuracy compared to the baseline network. We
believe this result demonstrates the difficulty of applying a
counterexample-guided training loop strategy for generating safe
networks compared an abstraction-guided methodology.

\subsection{Collision Detection Dataset}

Our second evaluation task focuses on the Collision Detection
Dataset~\cite{Ehlers17} where a neural network controller is used to
predict whether two vehicles running curve paths at different speeds
would collide. The network takes as input a feature vector of size
$6$, containing the information of distances, speeds, and directions
of the two vehicle. The network output prediction score are used to
classify the scenario as a colliding or non-colliding case.

A total of $500$ correctness properties are proposed in the Collision
Detection dataset that identify the safety margins around particular
data points. The network presented in the dataset respects $328$ such
properties. In our evaluation, we use a 3-layer fully-connected neural
network controller with 50, 128, 50 neurons in different hidden
layers. Using the same training configurations as in
Section~\ref{subsec:acas-results} and evaluating on the same training
and test sets provided in the dataset, the results are shown in
Table~\ref{tab:collision-numbers}.
\begin{table}[!t]
  \caption{Applying \name\ to Collision Detection Dataset.}
  \label{tab:collision-numbers}
  \centering
  \setlength{\tabcolsep}{3pt}
  \begin{tabular}{l *6c}
    \toprule
    {} & Refinement & Enforced & Accuracy & Time\\
    \midrule
    Original \cite{Ehlers17} & N/A & 328/500 & 99.87\% & N/A\\
    \addlinespace
    \name & \makecell{Yes\\No} & \makecell{481/500\\420/500} & \makecell{96.83\%\\86.3\%} & \makecell{583s\\419s}\\
    \bottomrule
  \end{tabular}
\end{table}
After $100$ epochs, \name converged to a local minimum and managed to
certify $481$ out of all $500$ safety properties. Although it did not
achieve zero correctness loss, ART can produce a solution that
satisfies significantly more correctness properties than the oracle
neural network, at the cost of only a small accuracy drop.

\section{Related Work}
\label{sec:related}

\para{Neural Network Verification.}
Inspired by the success of applying program analysis to large software
code bases, abstract interpretation-based techniques have been adapted
to reason about ANNs by developing efficient abstract transformers
that relax nonlinearity of activation functions into linear inequality
constraints~\cite{Ehlers17,SinghGMPV18,SG+18,DeepPoly,AI2,Gowal18,DiffAI}.
Similar approaches~\cite{zhang18,Weng18,Wang18a,Wang18b} encode
nonlinearity via linear outer bounds of activation functions and may
delegate the verification problem to SMT
solvers\cite{Reluplex,Marabou} or Mixed Integer Programming
solvers~\cite{cheng17,Dutta18,Tjeng19}.
Most of those verifiers focus on robustness properties only and do not
support verifiable training of network-wide correctness properties.
For example, \cite{DeepPoly} encodes concrete ANN operations into
ELINA~\cite{Elina}, a numeric abstract transformer, and therefore
disables opportunities for training or optimization thereafter.

Correctness properties may also be retrofitted onto a trained neural
network for safety concerns~\cite{ZX+19,AB+18,BK+15,realsyn}. These
approaches usually synthesize a reactive system that monitors the
potentially controller network and corrects any potentially unsafe
actions. Comparing to correct-by-construction methods, runtime
overheads are inevitable for such \emph{post facto} shielding
techniques.\\

\para{Correctness Properties in Neural Networks.}
There have been a large number of recent efforts that have explored
verifying the \emph{robustness} of networks against adversarial
attacks~\cite{Madry18,pei17,Goodfellow15}. Recent work has shown how
symbolic reasoning approaches~\cite{ReluVal,AI2} can be used to help
validate network robustness; other efforts combine optimization
techniques with symbolic reasoning to guide symbolic
analysis~\cite{AP+19}. Our approach looks at the problem of
verification and certification from the perspective of general safety
specifications that are typically richer than notions of robustness
governing these other techniques and provide the
correct-by-construction guarantee upon training termination.  Encoding
logical constraints other than robustness properties into loss
functions has been explored in
\cite{DL2,XuZFLB18,Minervini018,Hu16}. However, they operate only on
concrete sample instances and do not provide any
correct-by-construction guarantees.\\



\para{Training over Abstract Domains.}
The closest approach to our setting is the work in~\cite{DiffAI,BV20}.
They introduced geometric abstractions that bound activations as they
propagate through the network via abstract
interpretation. Importantly, since these convex abstractions are
differentiable, neural networks can optimize towards much tighter
bounds to improve the verified accuracy.
A simple bounding technique based on interval bound propagation was
also exploited in~\cite{Gowal18} (similar to the interval domain
from~\cite{DiffAI}) to train verifiably robust neural networks that
even beat the state-of-the-art networks in image classification tasks,
demonstrating that a correct-by-construction approach can indeed save
the need of more expensive verification procedures in challenging
domains.  They did not, however, consider verification in the context
of global safety properties as discussed here, in which the
over-approximation error becomes non-negligible; nor did they
formulate their approach to be parametric in the specific form of the
abstractions chosen.
Similar ideas have been exploited in provable defenses
works~\cite{WongK18,WongSMK18,RaghunathanSL18,BV20}, however, they
apply best-effort adversarial defenses only and provide no guarantee
upon training termination.

\section{Conclusions}
\label{sec:conc}

This paper presents a correct-by-construction toolchain that can train
neural networks with provable guarantees. The key idea is to optimize
a neural network over the abstraction of both the input space and the
network itself using abstraction refinement mechanisms. Experimental
results show that our technique realizes trustworthy neural network
systems for a variety of properties and benchmarks with only mild
impact on model accuracy.


\ifCLASSOPTIONcompsoc
  \section*{Acknowledgments}
\else
  \section*{Acknowledgment}
\fi

This work was supported by C-BRIC, one of six centers in JUMP, a
Semiconductor Research Corporation (SRC) program sponsored by DARPA;
NSF under award CCF-1846327; and NSF under Grant No. CCF-SHF 2007799.


\bibliographystyle{IEEEtran}
\bibliography{bib}

\appendix
\section{Appendix}

\subsection{Proofs}
\label{subsec:all-proofs}

\overapproxsoundness*
\begin{proof}
  Straightforward after unfolding Definition~\ref{def:absdom} and
  applying the properties of concrete and abstract transformers.
\end{proof}

\concretelosssoundness*
\begin{proof}
  Straightforward after unfolding Definition~\ref{def:conc-loss} and
  applying the fact that distance function $g(\cdot, \cdot) \ge 0$
  always holds.
\end{proof}

\zerodist*
\begin{proof}
  \label{prf:zero-dist}
  When $L_{\mathcal D}(F, \Phi_{\init}, \Phi_{\safe}) = 0$, by
  Definition~\ref{def:safety-dist},
  \[
    \max_{p \in \gamma(F_{\mathcal D}(\alpha(\Phi_{\init})))} dist(p,
    \Phi_{\safe}) = 0.
  \]
  Since $dist(\cdot)$ is a non-negative function, we have:
  \[
    \forall p \in \gamma(F_{\mathcal D}(\alpha(\Phi_{\init}))), ~ dist(p,
    \Phi_{\safe}) = 0.
  \]
  By Definition~\ref{def:absdom}, we have
  $\Phi_{\init} \subseteq \gamma(\alpha(\Phi_{\init}))$. By Theorem
  \ref{thm:approx-soundness}, we have
  \[
    \forall x, ~ x \models \Phi_{\init} \implies F(x) \in
    \gamma(F_{\mathcal D}(\alpha(\Phi_{\init}))).
  \]
  Hence,
  \[
    \forall x, ~ x \models \Phi_{\init} \implies dist(F(x), \Phi_{\safe}) =
    0.
  \]
  By Theorem~\ref{thm:concrete-loss-soundness}, it means
  \[
    \forall x, ~ x \models \Phi_{\init} \implies F(x) \models \Phi_{\safe}.
  \]
  Thereby proved
  $L_{\mathcal D}(F, \Phi_{\init}, \Phi_{\safe}) = 0 \implies F \models
  (\Phi_{\init}, \Phi_{\safe})$.
\end{proof}

\sufficientviarefine*
\begin{proof}
  By induction on Definition \ref{def:splitting-refinement},
  \begin{itemize}
  \item When $S_1 = S_2$, obviously $F \models S_2$;

  \item When
    $S_2 = \left\{ \Phi = \left(\Phi_{\init}, \Phi_{\safe}\right)
    \right\}$:

    By Definition \ref{def:splitting}, from $F \models S_1$ we have
    \[
      \bigwedge_{(\Psi_{\init}, \Psi_{\safe}) \in S_1} F \models
      (\Psi_{\init}, \Psi_{\safe})
    \]
    From
    \[
      \bigwedge_{(\_, \Psi_{\safe}) \in S_1} \Psi_{\safe} =
      \Phi_{\safe}
    \]
    we have
    \[
      \bigwedge_{(\Psi_{\init}, \Psi_{\safe}) \in S_1} F \models
      (\Psi_{\init}, \Phi_{\safe})
    \]
    Now that
    \[
      \Phi_{\init} = \bigcup_{(\Psi_{\init}, \_) \in S_1}
      \Psi_{\init},
    \]
    by Definition \ref{def:splitting}, we have
    $F \models (\Phi_{\init}, \Phi_{\safe})$. Thus $F \models S_2$.

  \item For transitivity rule, by induction hypothesis.
  \item For composition rule, by induction hypothesis.
  \end{itemize}
  All cases proved.
\end{proof}

\zerosafeloss*
\begin{proof}
  \label{prf:zero-safe-loss}
  Unfold the definitions, the proof is straightforward after applying
  Theorem~\ref{thm:zero-dist} and the fact that all abstract
  correctness losses are non-negative.
\end{proof}

\begin{lemma}[Valid Refinement]
  \label{lem:valid-refinement}
  For any input space abstraction $S$, the code snippet of
  Fig.~\ref{alg:vegal} starting from Line~\ref{alg:vegal:refine-start}
  to Line~\ref{alg:vegal:refine-end} yields an input space abstraction
  $S'$ such that $S' \sqsubseteq S$.
\end{lemma}
\begin{proof}
  \label{prf:valid-refinement}
  In the code snippet, original input space abstraction is divided
  into two parts, $T$ and $S'$. $S'$ remains the same throughout
  execution, so $S' \sqsubseteq S'$.

  For each correctness property
  $\left(\Phi_{\init}, \Phi_{\safe} \right)$ in $T$, the Refine
  procedure in Fig.~\ref{alg:vegal} is called to generate two new
  input predicates $\Phi_{\init}^1$ and $\Phi_{\init}^2$. It is easy
  to show that $\Phi_{\init}^1 \cup \Phi_{\init}^2 = \Phi_{\init}$. So
  every new pair
  \[
    \left\{ \left( \Phi_{\init}^1, \Phi_{\safe} \right), \left(
        \Phi_{\init}^2, \Phi_{\safe} \right) \right\} \sqsubseteq \left\{
      \left(\Phi_{\init}, \Phi_{\safe} \right) \right\}.
  \]
  By composition rule of Definition~\ref{def:splitting-refinement},
  the union of $S'$ and every such new pair refines original $S$.
\end{proof}

\artsoundness*
\begin{proof}
  From Lemma \ref{lem:valid-refinement}, we know for any input space
  abstraction $S'$ generated during the execution of
  Fig.~\ref{alg:vegal}, $S' \sqsubseteq S$. Then by Theorem
  \ref{thm:zero-safe-loss} and Theorem
  \ref{thm:sufficient-via-refine}, we have
  \[
    L_{\mathcal D}(F, S') = 0 \implies F \models S' \implies F \models
    S.
  \]
\end{proof}

\end{document}